\documentclass[10pt,twocolumn,letterpaper]{article}
\usepackage[pagenumbers]{iccv}
\usepackage{enumitem}
\usepackage{nicefrac}
\usepackage{times}
\usepackage{epsfig}
\usepackage{graphicx}
\usepackage{amsmath}
\usepackage{amssymb}
\usepackage{bm}
\usepackage{xspace}
\usepackage{microtype}
\usepackage{mathalfa}
\usepackage{graphicx}
\usepackage{graphbox}
\usepackage{array}
\usepackage{enumitem}
\usepackage{animate}
\usepackage{xspace} 
\usepackage{ctable}
\usepackage{multirow}
\usepackage{caption}
\usepackage{pifont}

\usepackage{booktabs}
\usepackage{colortbl}

\newcommand{\xmark}{\ding{55}}%

\makeatletter
\renewcommand{\paragraph}{\@startsection{paragraph}{4}{\z@}{1.2ex plus   
0.5ex minus .2ex}{-1em}{\normalsize\bf}}
\makeatother

\newlength\savewidth

\definecolor{baselinecolor}{gray}{.9}

\newcommand{\graymidrule}{\arrayrulecolor{gray!50}\midrule\arrayrulecolor{black}}

\definecolor{iccvblue}{rgb}{0.21,0.49,0.74}
\definecolor{LightCyan}{rgb}{0.88,1,1}
\definecolor{LightRed}{rgb}{1,0.5,0.5}
\definecolor{LightYellow}{rgb}{1,1,0.88}
\definecolor{Grey}{rgb}{0.75,0.75,0.75}
\definecolor{DarkGrey}{rgb}{0.55,0.55,0.55}
\definecolor{LightGreen}{rgb}{0.0, 0.70, 0.0}
\usepackage[pagebackref,breaklinks,colorlinks,allcolors=iccvblue]{hyperref}
\def\paperID{11595} 
\def\confName{ICCV}
\def\confYear{2025}

\title{LazyMAR: Accelerating Masked Autoregressive Models via Feature Caching}

\author{
Feihong Yan$^{1}$\thanks{Equal contribution. $^\dag$Corresponding authors.} \quad 
Qingyan Wei$^{2*}$ \quad 
Jiayi Tang$^{3}$ \quad 
Jiajun Li$^{4}$ \\
Yulin Wang$^{5}$ \quad 
Xuming Hu$^{6}$ \quad 
Huiqi Li$^{1\dagger}$ \quad 
Linfeng Zhang$^{7\dagger}$ \\[0.5em]
\hspace{-2em}
$^1$Beijing Institute of Technology \ \ \ 
$^2$Central South University \ \ \ 
$^3$China University of Mining and Technology \\
\hspace{-2em}
$^4$University of Electronic Science and Technology of China \ \ \ 
$^5$Tsinghua University \\
\hspace{-2em}
$^6$Hong Kong University of Science and Technology(Guangzhou) \ \ \  
$^7$Shanghai Jiaotong University \\\vspace{-1em}
}

\begin{document}
\maketitle
\begin{abstract}

Masked Autoregressive (MAR) models have emerged as a promising approach in image generation, expected to surpass traditional autoregressive models in computational efficiency by leveraging the capability of parallel decoding.
However, their dependence on bidirectional self-attention inherently conflicts with conventional KV caching mechanisms, creating unexpected computational bottlenecks that undermine their expected efficiency. To address this problem, this paper studies the caching mechanism for MAR by leveraging two types of redundancy:
\textbf{Token Redundancy} indicates that a large portion of tokens have very similar representations in the adjacent decoding steps, which allows us to first cache them in previous steps and then reuse them in the later steps. \textbf{Condition Redundancy} indicates that the difference between conditional and unconditional output in classifier-free guidance exhibits very similar values in adjacent steps. Based on these two redundancies, we propose LazyMAR, which introduces two caching mechanisms to handle them one by one.
LazyMAR is training-free and plug-and-play for all MAR models.
Experimental results demonstrate that our method achieves 2.83$\times$ acceleration with almost no drop in generation quality.  Our codes will be released in   \href{https://github.com/feihongyan1/LazyMAR}{\texttt{\textcolor{cyan}{https://github.com/feihongyan1/LazyMAR}}}.

\end{abstract}    
\section{Introduction}
\label{sec: intro}

Autoregressive models have shown exceptional capabilities not only in natural language processing~\cite{brown2020language,radford2018improving} but also in image generation~\cite{van2016pixel,chen2020generative,yuscaling,tian2025visual}. However, their token-by-token generation paradigm imposes strict sequential dependencies between image tokens, leading to less efficiency. To overcome this limitation, Masked Autoregressive (MAR) models~\cite{chang2022maskgit,li2023mage,li2025autoregressive} were introduced to enable parallel token prediction to improve computation efficiency while maintaining and even achieving better quality.

However, existing MARs still suffer from unpleasant computation efficiency despite their abilities of parallel decoding, which stems from their reliance on bidirectional attention mechanisms. The casual attention in autoregressive models allows the utilization of KV cache, which stores the $K$ matrix and $V$ matrix of tokens computed in previous steps and loads them during the following computation. Since the bidirectional attention in MARs requires simultaneous access to all tokens, it prevents the usage of KV cache and thereby reduces the efficiency of MARs. This limitation necessitates a novel caching mechanism tailored to MARs.

To address this issue, we begin by analyzing the computation redundancy in MARs, including token redundancy and condition redundancy. Then, we introduce LazyMAR which accelerates MAR by reducing the two redundancies.
\begin{figure}[t]
    \includegraphics[width=\linewidth]{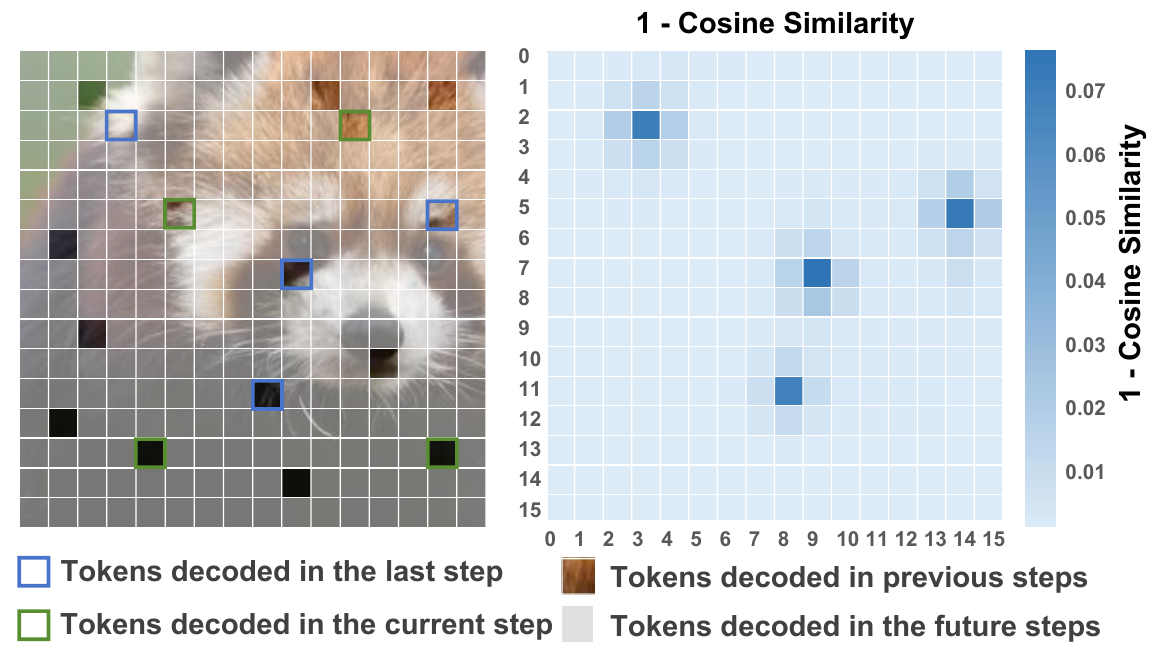}%
    \caption{
    \small 
    \textbf{Token Redundancy in MARs}: In each decoding step of MAR, tokens can be divided into four types, including the token to be decoded in this step ($t$), the token just decoded in the last step ($t-1$), the tokens that have decoded before last step, and the tokens that have not been decoded. Different tokens exhibit different similarities in the adjacent decoding steps.
    }%
    \vspace{-0.2cm}\label{fig:token_cache}%
\end{figure}%

\paragraph{Token Redundancy}: As shown in Figure~\ref{fig:token_cache}, the tokens in MARs can be divided into four groups based on their time of being decoded, including the tokens that decoded in the current step ($t$) and the last step ($t-1$), the tokens decoded in the steps before the last step ($\textless t-1$), and the future steps ($\textgreater t$). The heatmap shows the cosine distance between the representation of tokens in the current and the last step. It is observed that \emph{most tokens exhibit close distance between the adjacent two steps, except for the tokens that have been just decoded in the last step.}

\noindent\textbf{Solution - Token Cache}: Based on token redundancy, we introduce \emph{token cache}, which aims to store the features of tokens in previous steps and then reuse them in the following steps for acceleration. Notably, to reduce the negative influence of using the token cache on the generation quality, we still perform computations for the first several decoding steps, which are essential to determine the basic content of images. Meanwhile, we initialize the token cache by storing the features computed in these steps.
Then, in the following steps, we compute all the tokens in the first three layers of the neural networks and then skip the computation of tokens that exhibit high cosine similarity with
their values in the previous steps by reusing their features in the token cache, while still computing the tokens that show a lower similarity to maintain the generation quality. Experimental results demonstrate that the token cache can skip around 84\% tokens without harming generation quality, on average.

\begin{figure}[t]
    \captionsetup{type=figure}%
    \includegraphics[width=\linewidth]{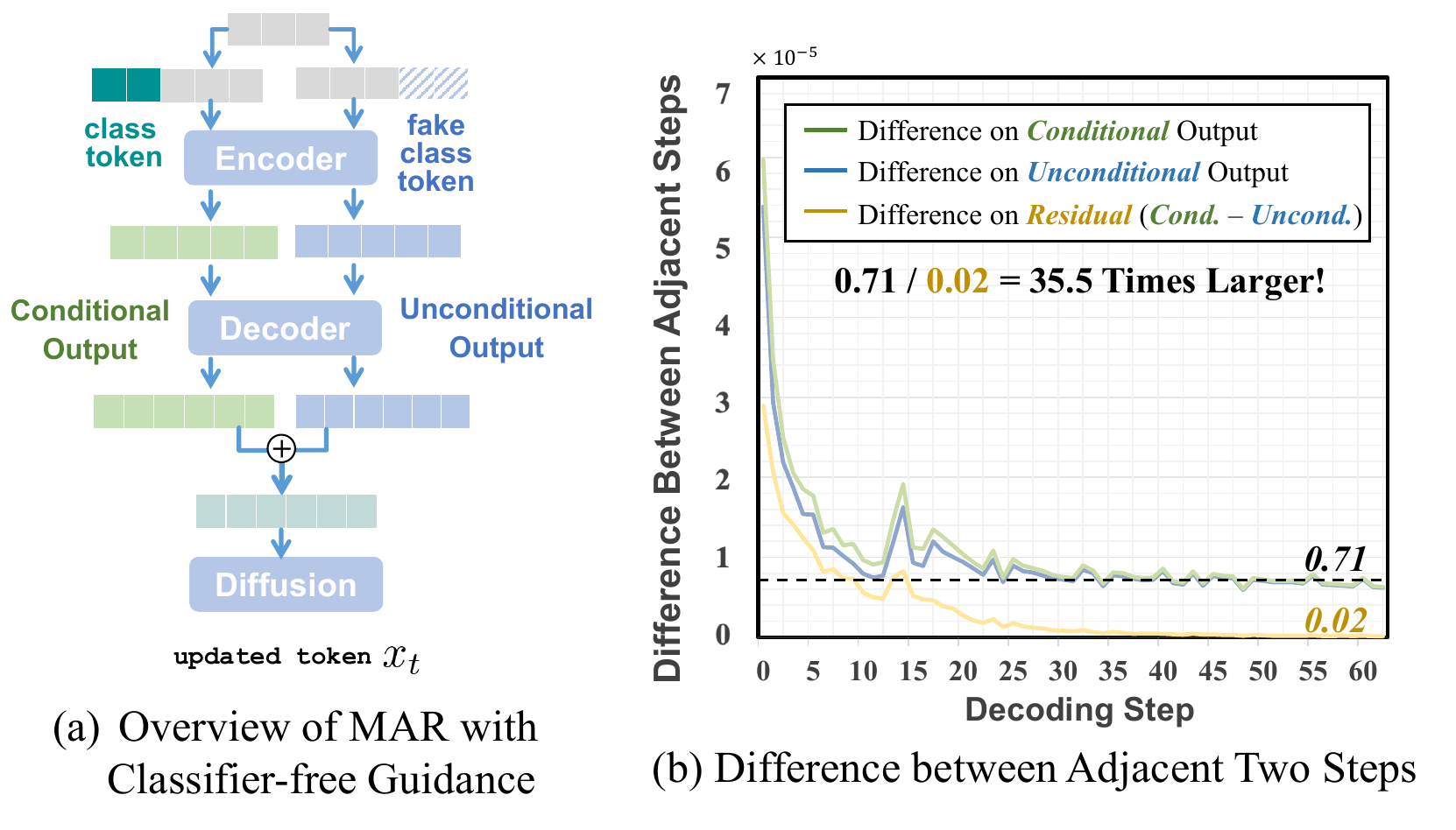}%
    \caption{
   \textbf{Conditional Redundancy in MARs:} Both the conditional and unconditional output exhibit significant distance in the adjacent decoding steps while their residual (cond. - uncond.) exhibits minor distance, which allows for caching and then reusing.  
    }%
    \label{fig:cfg_principle}%
\end{figure}%
\paragraph{Condition Redundancy}: The classifier-free guidance has been widely utilized in MARs to control the content of generated images via two parallel inputs, including one with conditional information and the other without it. The two inputs are both computed in MAR and their corresponding outputs and then mixed in a proper proportion. 
Obviously, such two-pathway paradigm doubles the computation costs in MAR.
Figure~\ref{fig:cfg_principle} shows the MSE difference of the conditional output, the unconditional output, and their residual (cond. - uncond.) between two adjacent steps. It is noted that both the difference of conditional output in adjacent steps and the unconditional output in adjacent steps exhibit significantly large values, while their residual demonstrates a minor value, which is around 35.5$\times$ smaller.

\noindent\textbf{Solution - Condition Cache}: The minor value of the difference on residual in the adjacent decoding steps indicates it is possible to cache the residual in the previous steps and then reuse them in the following steps. Concretely, we can skip the computation in the unconditional branch in the following steps by approximating it with the cached residual and the conditional output in this step. Obviously, skipping one of the two branches reduces the computation costs by 50\%.

\noindent\textbf{Periodic Cache-Reuse-Refreshment}: By leveraging the token redundancy and the conditional redundancy, the token cache and condition cache bring significant benefits in efficiency to MAR. However, the progressive accumulation of approximation errors through iterative cache reuse leads to exponential error amplification. To mitigate this problem, we implement a periodic ``cache-resue-refreshment strategy'': at scheduled intervals (every K decoding steps), we disable the caching mechanism, perform full computations of all the tokens and both the conditional and unconditional pathways, and refresh the cache with these ground-truth values.

Experimental results demonstrated that the combination of token cache and condition cache, which is referred to as LazyMAR, leads to around 2.83$\times$ acceleration on MAR-Diffusion with almost no drop in generation quality, outperforming directly reducing the sampling steps of MAR by 0.23 FID and 13.7 Inception Score. Besides, LazyMAR is plug-and-play for all MARs without requirements on training or searching.
In summary, our contributions are three-fold.
\begin{itemize}
    \item We introduce a framework of feature caching in MAR, including the periodic cache-reuse-refreshment strategy, the token cache, and the conditional cache.
    \item Motivated by the high similarity for tokens in adjacent decoding steps, we propose  \emph{token cache} for MAR to cache and then reuse the previously computed features while still computing the important and dissimilar tokens.
    \item Motivated by the minor difference between the residual from conditional and unconditional output across adjacent decoding steps, we propose \emph{condition cache} for MAR to cache and then reuse the previously computed residuals.
    \item Experimental results demonstrate that LazyMAR leads to 2.83 $\times$ acceleration with only 0.1 drop in FID. Besides, LazyMAR is plug-and-play for all MAR models without requirements of training.
    
\end{itemize}

\section{Related Work}
\subsection{(Masked) Autoregressive Models }

\paragraph{Autoregressive Models} 

Autoregressive language models~\cite{achiam2023gpt,brown2020language,ouyang2022training,radford2018improving,radford2019language} have demonstrated remarkable success in natural language processing by predicting tokens sequentially. This paradigm has been extended to visual modeling, where analogous approaches have been developed. Early works in autoregressive visual modeling~\cite{chen2020generative,van2016pixel,gregor2014deep,van2016conditional} focused on generating images pixel by pixel in a raster-scan order, but subsequent research shifted towards representing images as joint distributions of image tokens, which can be broadly categorized into two approaches: Autoregressive (AR) models~\cite{esser2021taming,yuvector,sun2024autoregressive,ramesh2021zero} and Masked Autoregressive (MAR) models~\cite{chang2022maskgit,li2023mage,li2025autoregressive,yu2025image}. Autoregressive models regard images as a sequence of discrete-valued tokens and rely on causal attention mechanisms, which can predict only one token per forward pass, limiting the model's efficiency significantly. LANTERN~\cite{jang2024lantern} and CoDe~\cite{chen2024collaborative} accelerate Autoregressive models through speculative decoding. PAR~\cite{wang2024parallelized} introduces parallelized autoregressive visual generation by grouping weakly dependent tokens for simultaneous prediction while maintaining sequential generation for strongly dependent ones. These methods aim to enable Autoregressive models with the ability to decode parallelly.

\paragraph{Masked Autoregressive Models} Masked Autoregressive (MAR) models leverage bi-directional attention to predict masked tokens simultaneously rather than sequentially. Notably, MaskGIT~\cite{chang2022maskgit} introduces a bidirectional transformer architecture for token modeling and proposes a parallel decoding strategy to enhance inference speed significantly. Building upon this foundation, MAGE~\cite{li2023mage} unifies generative modeling and representation learning via variable masking ratios and semantic tokens. MAR~\cite{li2025autoregressive} models per-token probability distributions using diffusion, enabling masked autoregressive modeling in continuous space without vector quantization, achieving new state-of-the-art results. However, the bi-directional attention in MARs makes it lose the ability of leverage KV Cache, which also poses challenges in inference efficiency, significantly limiting the scalability and inference speed of these models.

\subsection{Cache Mechanism for Visual Generation}
Large Language Models (LLMs) utilize key-value (KV) cache to optimize inference efficiency by storing and reusing key and value that are previously computed~\cite{zhang2023h2o,chen2024nacl}. Naturally, Visual Autoregressive models, employing analogous causal attention mechanisms, can benefit from KV caching for accelerated inference~\cite{sun2024autoregressive,tian2025visual,xie2024show}. Besides, Diffusion models have also adopted cache-based acceleration techniques. DeepCache~\cite{ma2024deepcache} accelerates the computation in UNet networks by caching upsampling block outputs across consecutive denoising steps. Similarly, Faster Diffusion~\cite{li2023faster} reuses encoder features from previous timesteps, omitting encoder computation selectively in adjacent steps. However, Masked Autoregressive models, being non-causal in nature, do not support KV caching. Moreover, existing diffusion caching techniques are specifically designed for U-Net architectures, making them incompatible with MARs, which typically employ Transformer architectures.

\section{Methodology}
\begin{figure*}[t!]
    \flushleft
    \includegraphics[width=\linewidth]{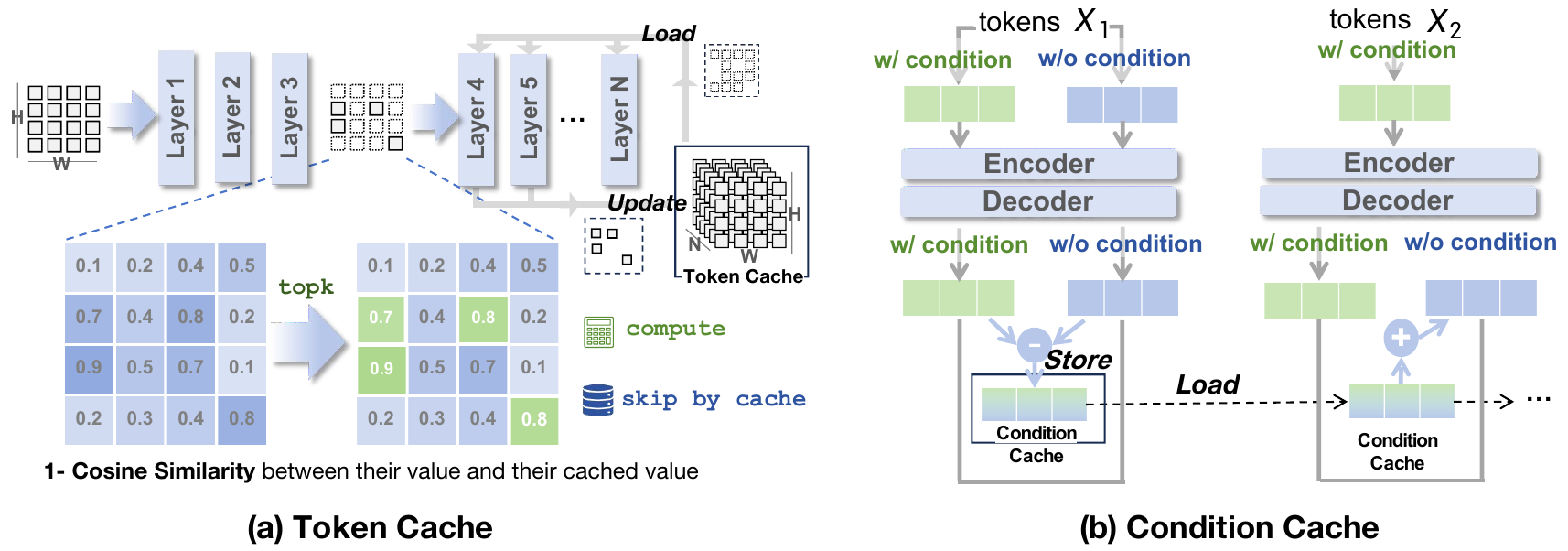}
    \caption{\small \textbf{The pipeline of LazyMAR:} We compute all the tokens and both the conditional and unconditional pathways in the first several steps, which are important to generate the basic structure of the image. 
    Then, we apply the token cache and condition cache periodically until the final step. \emph{(\expandafter{a}) Token Cache:} In the first step of each period, we compute all the tokens and store their features to initialize the two caches. 
    Then, in the following steps, we only compute all tokens in the first three layers and compare their difference with their values in previous steps. After that, we only compute the tokens with large differences in the following layers and skip other tokens by reusing their features in the token cache. 
    Meanwhile, we update the token cache with the features of tokens that have been computed. \emph{(\expandafter{b}) Condition Cache:} We compute both the conditional and unconditional pathways and store their residuals at the first step of each period. Then, we compute only the condition pathway and then approximate the output of the unconditional pathway by reusing the residual in the condition cache.
    }
    \label{fig: overview}
\end{figure*}

\subsection{Preliminary}

\paragraph{Autoregressive Models with KV Cache}
Traditional autoregressive models decompose the joint probability as:
$p(\mathbf{x}) = \prod_{i=1}^{n} p(x_{i} \mid \mathbf{x}_{<i})$
where 
$x_{\textless i}=[x_1,…,x_{i-1}]$.
Then, the attention computation at step $i$ is:
\begin{equation}
\text{Attention}(\mathbf{Q}_i, \mathbf{K}_{1:i}, \mathbf{V}_{1:i}) = \text{softmax}\left(\frac{\mathbf{Q}_i\mathbf{K}_{1:i}^\top}{\sqrt{d}}\right)\mathbf{V}_{1:i}
\end{equation}
where
$Q_i\in \mathbb{R}^{1\times d}$ is the query vector for the $i-$th token. 
$K_{1:i} \in \mathbb{R}^{i\times d}$  and $V_{1:i} \in \mathbb{R}^{i\times d}$ are the key and value matrix of first $i$ tokens, respectively.
The \textbf{KV Cache} stores historical key and value pairs $\{k_j, v_j\}_{j=1}^{i-1}$
  from previous steps, enabling incremental computation:
\begin{equation}
\mathbf{K}_{1:i} = [\underbrace{\mathbf{K}_{1:i-1}}_{\text{cached keys}}, \mathbf{k}_i], \quad
\mathbf{V}_{1:i} = [\underbrace{\mathbf{V}_{1:i-1}}_{\text{cached values}}, \mathbf{v}_i]
\end{equation}
which reduces computational complexity from $O(n 
^2)$ to $O(n)$ for n tokens.

\paragraph{MAR's Incompatibility with KV Cache}
Masked autoregressive models predict token subsets $\{\mathbf{X}^k\}_{k=1}^K$  through
$p(\mathbf{X}) = \prod_{k=1}^K p(\mathbf{X}^k \mid \mathbf{X}^{<k})$,
where $\mathbf{X}^{<k}$=$\cup_{j=1}^{k-1}\mathbf{X}^j$. Bidirectional attention allows all known tokens to interact:
\begin{equation}
\text{Attention}(\mathbf{Q}^k, \mathbf{K}^{\leq k}, \mathbf{V}^{\leq k}) = \text{softmax}\left(\frac{\mathbf{Q}^k (\mathbf{K}^{\leq k})^\top}{\sqrt{d}}\right)\mathbf{V}^{\leq k},
\end{equation}
which introduces the incompability for KV cache for the following two reasons.

\noindent \emph{Dynamic Updates}: Historical tokens' representations \emph{get updated} through attention layers denoted as $f$: \begin{equation}
\mathbf{k}_j^{(k)} = f(\mathbf{k}_j^{(k-1)}, \mathbf{X}^k), \quad
\mathbf{v}_j^{(k)} = g(\mathbf{v}_j^{(k-1)}, \mathbf{X}^k) \quad \forall j \in \mathbf{X}^{<k}
\end{equation}
\noindent \emph{Full Dependencies}: Each step $k$ requires \emph{recomputing all keys and values} due to: \begin{equation}
\mathbf{K}^{\leq k} = [\mathbf{K}^{<k}, \mathbf{K}^k], \quad \mathbf{K}^{<k} \neq \text{Cache}(\mathbf{K}^{<k-1})
\end{equation}
which prevents MAR from using KV cache and thus leads to significant inefficiency in computation.

\paragraph{Classifier-Free Guidance in MAR}
The classifier-free guidance (CFG) in MAR is utilized to control the generated content by 
combining conditional and unconditional pathways. It introduces a subset of conditional tokens carrying the conditional information (\emph{e.g.} language guidance, class guidance) to the image tokens in the conditional pathway, and a subset of corresponding fake conditional tokens whose values are zero in the unconditional pathways. For the images tokens as $\mathbf{X}$, this can be formulated as 
\begin{equation}
    \mathbf{X_{\text{cond}}}= \mathbf{X}\cup \{c_i\}_{i=1}^C, ~~\mathbf{X_{\text{uncond}}}= \mathbf{X}\cup \{\hat{c_i}\}_{i=1}^C,
\end{equation}
where ${c_i}$ indicates the conditional token and $C$ denotes its amount. ${\hat{c}_i}=\mathbf{0}$ denotes the fake conditional tokens. Then, by denoting MAR as $\mathcal{F}$, the conditional and unconditional output can be formulated as 
\begin{equation}
    \mathcal{F(\mathbf{X})} = \mathcal{F}(\mathbf{X}_{\text{cond}}) + \gamma \cdot \mathcal{F}(\mathbf{X}_{\text{uncond}}), 
\end{equation}
where $\gamma$ is a hyper-parameter to balance them. Obviously, the two pathways in CFG double the computation costs.
\subsection{LazyMAR}

\subsubsection{Token Cache}
\paragraph{Token Selection for Caching}
As shown in Figure~\ref{fig:token_cache}, most tokens in MAR exhabit very similar values in the adjacent decoding steps, making it possible to cache the features of tokens in the previous steps and then resue them in the following steps. However, such caching and reusing may harm the generation quality of MAR. To minimize its influence, we need to carefully analyze the influence from caching different kinds of tokens. Concretely, we find tokens can be divided into the following four types.
\begin{itemize} 
  \item[1] \textbf{Tokens decoded in previous steps} can still have influence to the other tokens but the generated content of them have been decided. Hence, the caching error introduces on them does not lead to significant harm to the generation results. Besides, since their content have been decided, their feature exhabit similar difference in adjacent steps. These two observations make caching these tokens does not harm model performance.
  \item[2] \textbf{Tokens decoded in the last step} shows very significant differences across the adjacent decoding steps since their value has been largely changed by the diffusion process during their decoding. These significant differences making it not suitable to cache them.
  \item[3] \textbf{Tokens decoded in the current step} directly decides the quality of the generated results. Thereby, caching on these tokens is significantly harmful to MAR.
  \item[4] \textbf{Tokens decoded in the future steps}
  exhibits differences in varying levels in the adjacent steps depending on their distances to the tokens decoded in the last step. 
  The different behaviors of these tokens indicates that they should be handled in different strategies.
\end{itemize}

The above analysis reveals that different tokens have different priorities for caching and computation. In this paper, we begin by designing two strategies as follows.

\begin{itemize}
    \item[1] \textbf{Observation-based Strategy} indicates that whether a token should be computed or reused from the cache is decided by the type of this token, as discussed above.
    \item[2]
    \textbf{Similarity-based Strategy} indicates that whether a token should be computed or reused from the cache is decided by their similarity compared with their cached values in the cache. 
    Intuitively, a higher similarity indicates that reusing this token leads to lower error, which encourages this token to be skipped by cache-reusing. 
    As shown in Figure~\ref{fig:relation_x_v} and discussed layer,  we find that there exists a strong positive correlation between the difference of tokens in the adjacent steps at the early layer (\emph{e.g.} 3rd layer) and the final layer, indicating that we can firstly forward the early layers without using the cache, then compute the difference between token features in adjacent steps,  and then skip the computation of the tokens with a smaller difference in the following layers. 
\end{itemize}
As shown in Table~\ref{tab:ablation_caching_strategies} and discussed layer, we find the similarity-based strategy shows a relatively better performance, and hence takes it as our default strategy.



\paragraph{Caching Pipeline}
Based on the above analysis and solution, now we formulate the proposed token cache as follows. 
For a period of steps from $i$ to $i+\tau$, we compute all the tokens and then store their features in the token cache at the first step $i$, which can be denoted as $\mathcal{C}_{\text{token}}^l:= \mathcal{F}^l(\mathbf{X}^i)$, where $:=$ indicates the operation of assigning value, $\mathcal{F}$ denotes the layers in MAR, and $l$ denotes the index of layers.
Then, in the following step $i+j\leq i+\tau$, we divide tokens into two sets, including one of both to be skipped by the cache, whose index can be denoted as $\mathcal{I}_{\text{cache}}$, and the other one to be fully computed, whose index set can be denoted as $\mathcal{I}_{\text{compute}}$. $\mathcal{I}_{\text{compute}}$ is decided by the similarity between the features of tokens in the early layers, which can be formulated as 
 \begin{equation}
\mathcal{I}_{~\text{Compute}}=\mathop{\arg\max }_{\{i_1, \dots, i_N\}\subseteq {\{1,\dots,n}\}} \{\mathcal{S}(x_{i_1}), \cdots, \mathcal{S}(x_{i_n})\},
\end{equation}
where $\mathcal{S}(x_i)=\nicefrac{\mathcal{F}^{l=3}(x_i)\cdot \mathcal{C}^{l=3}[x_i]}{\|\mathcal{F}^{l=3}(x_i)\cdot\| \|\mathcal{C}^{l=3}[x_i]\|}$ denotes the cosine similarity between the value of tokens in the current and the previous steps at the 3$_{rd}$ layer. $N$ denotes the number of tokens to be computed. $\mathcal{I}_{\text{cache}}$ is the complementary set of $\mathcal{I}_{\text{compute}}$.
Then, as shown in Figure~\ref{fig: overview}(a), the computation result of each layer can be formulated  as
\begin{equation}
\mathcal{C}_{\text{token}}[\mathcal{I}_{\text{cache}}] \cup \mathcal{F}(\mathbf{X}^{i+j}[\mathcal{I}_{\text{compute}}]),
\end{equation}
where the left term indicates results that are loaded from the cache, which takes no computation costs,  while the right term indicates the results from computations. By applying such caching to both self-attention and MLP layers, abundant computation can be skipped, leading to better efficiency.

\subsubsection{Condition Cache}
The classifier-free guidance in MAR introduces a conditional computation pathway and an unconditional computation pathway and thus doubles the computation overhead. As shown in Figure~\ref{fig:cfg_principle}, the residual from the conditional and unconditional output exhibits a very minor value in adjacent steps, providing the possibility of condition cache. By denoting the conditional cache as $\mathcal{C}_{\text{cond}}$, for a period of steps from $i$ to $i+\tau$, we compute both the conditional and unconditional pathways and store their residual at the first step $i$. As shown in Figure~\ref{fig: overview}(b), it can be formulated as 
\begin{equation}
    \mathcal{C}_{\text{cond}}:= \mathbf{X}^i_{\text{uncond}}-\mathbf{X}^i_{\text{cond}}.
\end{equation} Then, in the following step $i+j\leq i+\tau$, we compute the condition pathway, and then approximate the output of the unconditional pathway as $\mathbf{X}^i_{\text{uncond}} = \mathbf{X}^i_{\text{cond}} + \mathcal{C}_{\text{cond}}$. In this way, conditional cache enables MAR to save half computation costs.



\section{Experiment}
\subsection{Experiment Settings}

\paragraph{Implementation Details}We evaluate our method on three versions of the MAR~\cite{li2025autoregressive} models: MAR-B (208M), MAR-L (479M), and MAR-H (943M). The experiments are conducted on ImageNet~\cite{deng2009imagenet} class-conditional generation benchmark at a resolution of $256 \times 256$. The diffusion process follows the standard setting~\cite{dhariwal2021diffusion,ho2020denoising}, with 100 diffusion steps during inference. We use the KL-16 image tokenizer~\cite{rombach2022high}, which divides the image into 256 tokens. Additionally, following MAR~\cite{li2025autoregressive}, we prepend 64 cls tokens in the encoder input sequence. To assess the effectiveness of our method comprehensively, we evaluate our method in two different numbers of decoding steps: 32, and 64. For a total of 32 decoding steps, we start caching from 4th step and set $\tau = 5$, skipping around 250 tokens in each caching decoding step. For a total of 64 decoding steps, we start caching from 5th step and set $\tau = 9$, skipping around 270 tokens at each caching decoding step. All other configurations remain consistent with the default MAR settings. ALL experiments are conducted on two NVIDIA 3090 GPUs and one Inter 12900K CPU using PyTorch 2.2 with FP16.

\paragraph{Evaluation Metrics} We generate 50k images using the 1000 classes from ImageNet1K and evaluate them using FID~\cite{heusel2017gans} and IS~\cite{salimans2016improved} as primary quality metrics. We assess computational efficiency using FLOPs, CPU latency and GPU latency. Our evaluation protocol follows the guidelines provided by~\cite{dhariwal2021diffusion}, ensuring a fair and consistent comparison.

\begin{figure*}[t!]
    \flushleft
    \includegraphics[width=\linewidth]{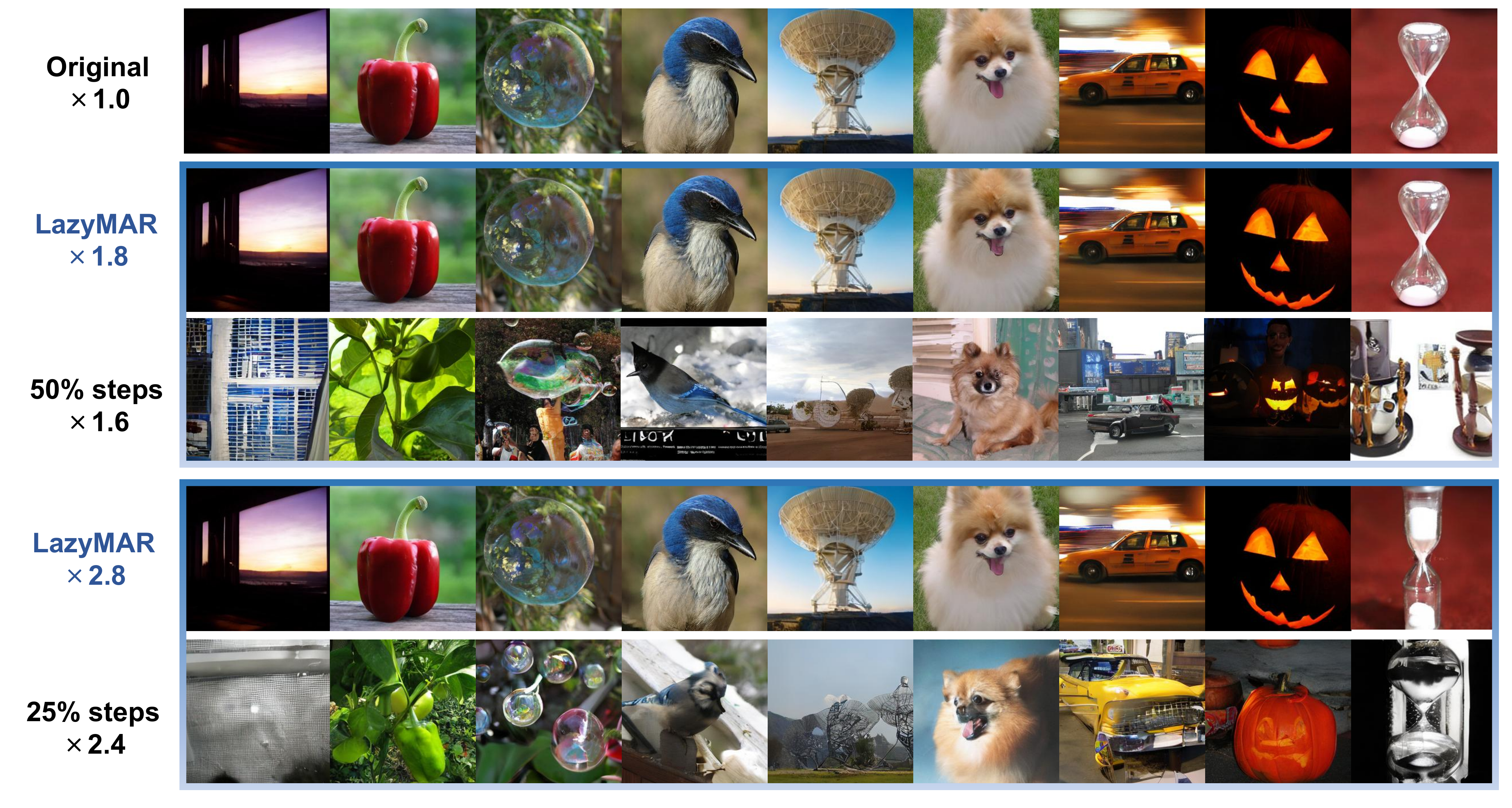}
    \caption{\small \textbf{Qualitative comparison} between LazyMAR and the acceleration achieved through step reduction. Experimental results show that our method maintains good consistency with the original images in both structure and details.
    }
    \label{fig: visualization}
\end{figure*}
\subsection{Qualitative Analysis}
We conducted a qualitative comparison between LazyMAR and the original MAR under different acceleration ratios. As shown in Figure~\ref{fig: visualization}, even with a higher acceleration ratio of 2.8$\times$, LazyMAR maintains high image quality and semantic fidelity. In most cases, the images generated by LazyMAR are almost indistinguishable from the original images. In contrast, acceleration through a reduction in decoding steps often results in distorted images, as one potential reason is that predicting too many tokens within a single decoding step impairs the coordination and perceptual alignment between the tokens decoded in the current step.

\begin{table*}[ht]
  \caption{
\textbf{Model comparison results} on ImageNet $256\times256$ class-conditional generation. ``MAR-B, -L, -H" denote MAR's base, large, and huge models. ``64, 32, 16" represent the total number of decoding steps.
}
  \centering
  \small
  \resizebox{\textwidth}{!}{\begin{tabular}{l | c c c c c | c c}
    \hline
    \toprule
    \multirow{2}{*}{\bf Method} & \multicolumn{5}{c|}{\bf Inference Efficiency} & \multicolumn{2}{c}{\bf Generation Quality}  \\
    \cmidrule(lr){2-6} \cmidrule(lr){7-8}
  & {\bf Latency(GPU/s)$\downarrow$} & {\bf Latency(CPU/s)$\downarrow$} & {\bf FLOPs(T)$\downarrow$} & {\bf Speed$\uparrow$} & {\bf Param} & {\bf FID $\downarrow$} & {\bf IS $\uparrow$} \\
  \midrule
  MaskGIT~\cite{chang2022maskgit} & 0.13 & 9.72 & - & 1.00 & 227M & 6.18 & 182.1 \\
  MAGE~\cite{li2023mage} & 1.60 & 12.60 & 4.19 & 1.00 & 307M & 6.93 & 195.8 \\
  LDM-4~\cite{esser2021taming} & 5.35 & 27.25 & 69.5 & 1.00 & 400M & 3.60 & 247.7 \\
  DiT-XL/2~\cite{Peebles2023} & 4.84 & 196.88 & 114.38 & 1.00 & 675M & 2.27 & 278.2 \\
  LlamaGen-XXL~\cite{sun2024autoregressive} & 1.98 & 736.49 & 3.19 & 1.00 & 1.4B & 3.09 & 253.6 \\
  LlamaGen-3B~\cite{sun2024autoregressive} & 2.07 & 1524.63 & 7.01 & 1.00 & 3.1B & 3.05 & 222.3 \\

  \midrule
  \textcolor{gray}{MAR-B, 64} &  \textcolor{gray}{0.47} &  \textcolor{gray}{28.97} &  \textcolor{gray}{15.49} &  \textcolor{gray}{1.00} &  \textcolor{gray}{208M} &  \textcolor{gray}{2.32} &  \textcolor{gray}{281.1} \\
  \graymidrule
  MAR-B, 32 & 0.29 & 18.11 & 9.54 & 1.62 & 208M & 2.47 & 273.1 \\
  LazyMAR-B, 64 & $0.21_{\textcolor{LightGreen}{-0.08}}$ & $11.08_{\textcolor{LightGreen}{-7.03}}$ & $5.54_{\textcolor{LightGreen}{-4.00}}$ & $2.80_{\textcolor{LightGreen}{+1.18}}$ & 208M & $\textbf{2.45}_{\textcolor{LightGreen}{-0.02}}$ & $\textbf{281.3}_{\textcolor{LightGreen}{+8.2}}$ \\
  \graymidrule
  MAR-B, 16 & 0.20 & 12.12 & 6.51 & 2.38 & 208M & 4.10 & 247.5 \\
  LazyMAR-B, 32 & $0.16_{\textcolor{LightGreen}{-0.04}}$ & $8.23_{\textcolor{LightGreen}{-3.89}}$ & $4.10_{\textcolor{LightGreen}{-2.41}}$ & $3.78_{\textcolor{LightGreen}{+1.40}}$ & 208M & $2.64_{\textcolor{LightGreen}{-1.46}}$ & $276.0_{\textcolor{LightGreen}{+28.5}}$ \\
  
  \midrule
  \textcolor{gray}{MAR-L, 64} &  \textcolor{gray}{0.93} &  \textcolor{gray}{59.66} &  \textcolor{gray}{35.05} &  \textcolor{gray}{1.00} &  \textcolor{gray}{479M} &  \textcolor{gray}{1.82} &  \textcolor{gray}{296.1} \\
  \graymidrule
  MAR-L, 32 & 0.57 & 36.25 & 21.15 & 1.66 & 479M & 2.05 & 281.7 \\
  LazyMAR-L, 64 & $0.40_{\textcolor{LightGreen}{-0.17}}$ & $22.82_{\textcolor{LightGreen}{-13.43}}$ & $12.52_{\textcolor{LightGreen}{-8.63}}$ & $2.80_{\textcolor{LightGreen}{+1.14}}$ & 479M & $\textbf{1.93}_{\textcolor{LightGreen}{-0.12}}$ & $\textbf{297.4}_{\textcolor{LightGreen}{+15.7}}$ \\
  \graymidrule
  MAR-L, 16 & 0.38 & 23.78 & 14.20 & 2.47 & 479M & 4.32 & 247.4 \\
  LazyMAR-L, 32 & $0.30_{\textcolor{LightGreen}{-0.08}}$ & $17.07_{\textcolor{LightGreen}{-6.71}}$ & $9.41_{\textcolor{LightGreen}{-4.79}}$ & $3.72_{\textcolor{LightGreen}{+1.25}}$ & 479M & $2.11_{\textcolor{LightGreen}{-2.21}}$ & $284.4_{\textcolor{LightGreen}{+37.0}}$ \\
  \midrule

  \textcolor{gray}{MAR-H, 64} &  \textcolor{gray}{1.74} &  \textcolor{gray}{116.61} &  \textcolor{gray}{69.06} &  \textcolor{gray}{1.00} &  \textcolor{gray}{943M} &  \textcolor{gray}{1.59} &  \textcolor{gray}{299.1} \\
  \graymidrule

  MAR-H, 32 & 1.04 & 70.97 & 42.12 & 1.64 & 943M & 1.92 & 285.5 \\
  LazyMAR-H, 64 & $0.75_{\textcolor{LightGreen}{-0.29}}$ & $43.12_{\textcolor{LightGreen}{-27.85}}$ & $24.38_{\textcolor{LightGreen}{-17.74}}$ & $2.83_{\textcolor{LightGreen}{+1.19}}$ & 943M & $\textbf{1.69}_{\textcolor{LightGreen}{-0.23}}$ & $\textbf{299.2}_{\textcolor{LightGreen}{+13.7}}$ \\ 
  \graymidrule
  MAR-H, 16 & 0.70 & 45.71 & 28.67 & 2.41 & 943M & 4.49 & 242.9 \\
  LazyMAR-H, 32 & $0.55_{\textcolor{LightGreen}{-0.15}}$ & $33.50_{\textcolor{LightGreen}{-12.21}}$ & $19.03_{\textcolor{LightGreen}{-9.64}}$ & $3.63_{\textcolor{LightGreen}{+1.22}}$ & 943M & $1.94_{\textcolor{LightGreen}{-2.55}}$ & $284.1_{\textcolor{LightGreen}{+41.2}}$ \\
  \bottomrule
  \hline
  \end{tabular}}
\vspace{-2mm}
\label{tab:main_table}
\end{table*}

\subsection{Quantitative Evaluations}
Table~\ref{tab:main_table} shows the quantitative results of our method. It is observed that: (1) Compared with the original MAR with similar GPU/CPU latency, our method achieves superior FID and IS, as well as better speedup ratios across all experimental configurations.
(2) Compared with the original MAR with the same model scale and decoding steps, we achieve about 2.8$\times$ speedup while maintaining comparable image generation quality. For example, LazyMAR-H, 64 achieves 2.83$\times$ speedup and 0.1 improvement in IS score, with only 0.1 increase in FID score compared with the best-performing setting of the original MAR (MAR-H, 64).
In contrast, MAR-H, 32, which has a lower speedup ratio compared with LazyMAR-H, 64, results in a 0.33 increase in FID and a 13.6 decrease in IS.
(3)
Additionally, LazyMAR-H, 64 and LazyMAR-L, 64 outperform their smaller counterparts (MAR-L, 64 and MAR-B, 64) in both computational efficiency and image generation quality. 
In summary, our method achieves great efficiency in inference while maintaining high-quality in the generation results.

\subsection{Ablation Study}

\begin{table*}[t]
  \caption{\textbf{Ablation studies} of the two cache strategies in our method on $256 \times 256$ class-conditional generation. The experimental results demonstrate the effectiveness of both caching mechanisms.}
  \centering
  \small
  \resizebox{\textwidth}{!}{\begin{tabular}{cc|cccccc}
  \toprule
  {\bf Token Cache} &  {\bf Condition Cache} &  {\bf Latency(GPU/s)$\downarrow$} &  {\bf Latency(CPU/s)$\downarrow$} &  {\bf FLOPs(T)$\downarrow$} &  {\bf Speed$\uparrow$} & {\bf FID$\downarrow$} &  {\bf IS$\uparrow$} \\
  \midrule
  \xmark & \xmark & 1.74 & 116.61 & 69.06 & 1.00 & 1.59 & 299.1 \\
  \midrule
  \checkmark & \xmark & $1.23_{\textcolor{LightGreen}{-0.51}}$ & $75.87_{\textcolor{LightGreen}{-40.74}}$ & $44.21_{\textcolor{LightGreen}{-24.85}}$ & $1.59_{\textcolor{LightGreen}{+0.59}}$ & $\textbf{1.62}_{\textcolor{DarkGrey}{+0.03}}$ & $297.9_{\textcolor{DarkGrey}{-1.2}}$ \\
  \xmark & \checkmark & $0.98_{\textcolor{LightGreen}{-0.76}}$ & $63.24_{\textcolor{LightGreen}{-53.37}}$ & $36.80_{\textcolor{LightGreen}{-32.26}}$ & $1.87_{\textcolor{LightGreen}{+0.87}}$ & $1.67_{\textcolor{DarkGrey}{+0.08}}$ & $\textbf{302.6}_{\textcolor{LightGreen}{+3.5}}$ \\
  
  \checkmark & \checkmark & $\textbf{0.75}_{\textcolor{LightGreen}{-0.99}}$ & $\textbf{43.12}_{\textcolor{LightGreen}{-73.49}}$ & $\textbf{24.38}_{\textcolor{LightGreen}{-44.68}}$ & $\textbf{2.83}_{\textcolor{LightGreen}{+1.83}}$ & $1.69_{\textcolor{DarkGrey}{+0.10}}$ & $299.2_{\textcolor{LightGreen}{+0.1}}$ \\
  \bottomrule
  \end{tabular}}
  \label{tab:ablation_cfg_token}
\end{table*}

\paragraph{Effectiveness of Two Caching Mechanisms} 
Compared with the standard MAR inference process, two caching mechanisms are introduced in the proposed method. The effectiveness of these two caching mechanisms is demonstrated in Table~\ref{tab:ablation_cfg_token}. The experimental results reveal two key findings: (1) Each caching mechanism independently contributes to significant reductions in computational overhead. (2) When both caching mechanisms are employed simultaneously, our method achieves optimal acceleration while simultaneously preserving the high fidelity and quality of generated images.

\begin{table*}[t]
  \caption{\textbf{Ablation studies} of six token selection strategies on ImageNet class-to-image generation.}
  \vspace{-0.1cm}
  \centering
  \resizebox{\textwidth}{!}{\begin{tabular}{l|cccccc}
  \toprule
  {\bf Method}  &  {\bf Latency(GPU/s)$\downarrow$} &  {\bf Latency(CPU/s)$\downarrow$} &  {\bf FLOPs(T)$\downarrow$} & {\bf Speed$\uparrow$} & {\bf FID$\downarrow$} &  {\bf IS$\uparrow$} \\
  \midrule
   \textcolor{DarkGrey}{Origin} & \textcolor{DarkGrey}{1.74} &  \textcolor{DarkGrey}{116.61} &  \textcolor{DarkGrey}{69.06} &  \textcolor{DarkGrey}{1.00} &  \textcolor{DarkGrey}{1.59} &  \textcolor{DarkGrey}{299.1} \\
   
  Condition Cache & $0.98_{\textcolor{LightGreen}{-0.76}}$ & $63.24_{\textcolor{LightGreen}{-53.37}}$ & $36.80_{\textcolor{LightGreen}{-32.26}}$ & $1.87_{\textcolor{LightGreen}{+0.87}}$ & $1.67_{\textcolor{DarkGrey}{+0.08}}$ & $302.6_{\textcolor{LightGreen}{+3.5}}$ \\
 \midrule
  ~~+\textbf{Unpredicted Token (\uppercase\expandafter{\romannumeral1})} & $0.75_{\textcolor{LightGreen}{-0.99}}$ & 
  $40.26_{\textcolor{LightGreen}{-76.35}}$ & 
  $22.60_{\textcolor{LightGreen}{-46.46}}$ & 
  $3.06_{\textcolor{LightGreen}{+2.06}}$ & 
  $3.61_{\textcolor{DarkGrey}{+2.02}}$ & 
  $248.1_{\textcolor{DarkGrey}{-51.0}}$ \\
  
  ~~+\textbf{Predicted Token (\uppercase\expandafter{\romannumeral2})} & $0.79_{\textcolor{LightGreen}{-0.95}}$ & 
  $43.16_{\textcolor{LightGreen}{-73.45}}$ & 
  $24.94_{\textcolor{LightGreen}{-44.12}}$ & 
  $2.77_{\textcolor{LightGreen}{+1.77}}$ & 
  $\textbf{1.79}_{\textcolor{DarkGrey}{+0.20}}$ & 
  $\textbf{291.5}_{\textcolor{DarkGrey}{-7.6}}$ \\
  
  \midrule
  
  ~~+\textbf{Random (\uppercase\expandafter{\romannumeral3})} & $0.75_{\textcolor{LightGreen}{-0.99}}$ & 
  $43.12_{\textcolor{LightGreen}{-73.49}}$ & 
  $24.38_{\textcolor{LightGreen}{-44.68}}$ & 
  $2.83_{\textcolor{LightGreen}{+1.83}}$ & 
  $1.95_{\textcolor{DarkGrey}{+0.36}}$ & 
  $288.3_{\textcolor{DarkGrey}{-10.8}}$ \\
  
  ~~+\textbf{Min-Similarity (\uppercase\expandafter{\romannumeral4})} & $0.75_{\textcolor{LightGreen}{-0.99}}$ & 
  $43.12_{\textcolor{LightGreen}{-73.49}}$ & 
  $24.38_{\textcolor{LightGreen}{-44.68}}$ & 
  $2.83_{\textcolor{LightGreen}{+1.83}}$ & 
  $2.10_{\textcolor{DarkGrey}{+0.51}}$ & 
  $283.4_{\textcolor{DarkGrey}{-15.7}}$ \\

  ~~+\textbf{LazyMAR (\uppercase\expandafter{\romannumeral5})} & $0.75_{\textcolor{LightGreen}{-0.99}}$ & 
  $43.12_{\textcolor{LightGreen}{-73.49}}$ & 
  $24.38_{\textcolor{LightGreen}{-44.68}}$ & 
  $2.83_{\textcolor{LightGreen}{+1.83}}$ & 
  
  $\textbf{1.69}_{\textcolor{DarkGrey}{+0.10}}$ & 
  $\textbf{299.2}_{\textcolor{LightGreen}{+0.1}}$ \\
  
  \bottomrule
  \end{tabular}}%
  \label{tab:ablation_caching_strategies}
\end{table*}

\paragraph{Token Selection Strategies} 
In this section, we study how different token selection strategies influence the quality of image generation. The following five token selection strategies are considered:

\begin{itemize}[label={}]
    \item (\textbf{\uppercase\expandafter{\romannumeral1}) Unpredicted Token:} caching the unpredicted tokens. 
    \item (\textbf{\uppercase\expandafter{\romannumeral2}) Predicted Token:} only caching the predicted tokens.
   \item  (\textbf{\uppercase\expandafter{\romannumeral3}) Random:} selecting tokens randomly for caching. 
    \item (\textbf{\uppercase\expandafter{\romannumeral4}) Min-Similarity:} caching the tokens with the minimum cosine similarity between token features in the adjacent decoding steps at the 3rd layer. 
    \item  (\textbf{\uppercase\expandafter{\romannumeral5}) LazyMAR:} caching the tokens with the maximum cosine similarity between token features in the adjacent decoding steps at the 3rd layer.  
\end{itemize}


Experimental results of the five strategies on ImageNet are provided in Table~\ref{tab:ablation_caching_strategies}. It can be observed that:
(1) Strategies \textbf{\uppercase\expandafter{\romannumeral1}} and \textbf{\uppercase\expandafter{\romannumeral4}} perform worst, indicating that caching the tokens that have not been decoded or those with the minimum cosine similarity will lead to non-negligible errors.
(2) Strategy \textbf{\uppercase\expandafter{\romannumeral2}} outperforms Strategy \textbf{\uppercase\expandafter{\romannumeral1}}, indicating that the decoded tokens obtain more stable semantics through the diffusion-based process, making them more suitable for feature reuse compared with unpredicted tokens.
(3) Our method (Strategy \textbf{\uppercase\expandafter{\romannumeral5}}) performs best, indicating that if a token shows high similarity with its corresponding cached features in early layers, its output will tend to maintain this high similarity. Caching these high-similarity tokens minimizes the impact on image generation quality effectively.

\begin{figure*}[t!]
    \flushleft
    \includegraphics[width=\linewidth]{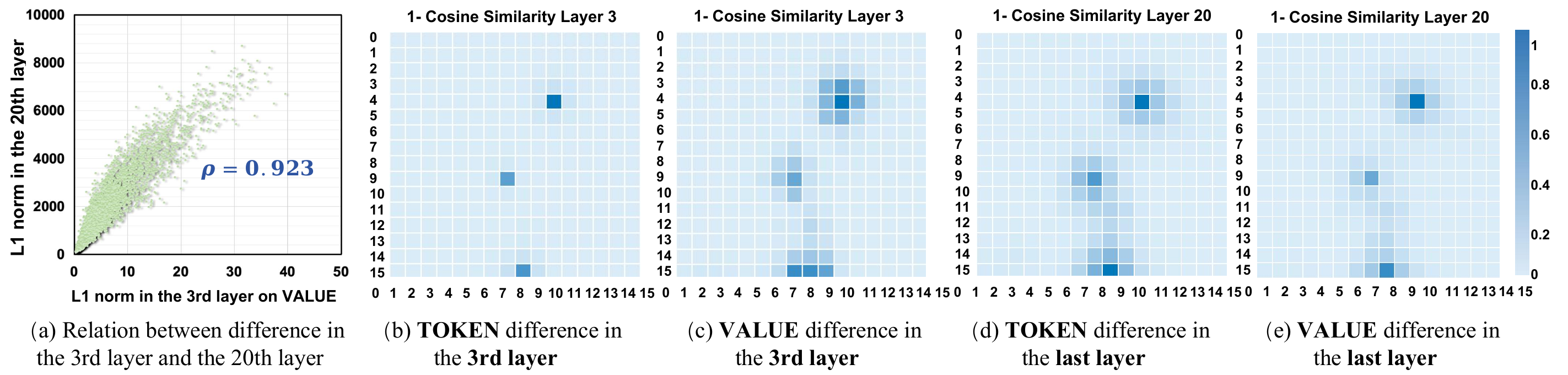}
    \vspace{-0.25cm}
    \caption{\small (a) \textbf{The relationship} between the difference of tokens in adjacent steps at the early layer and the final layer. A strong positive correlation is observed, with a Pearson correlation coefficient of $\rho = 0.923$. (b)-(e) \textbf{show the difference} between the cached features and the corresponding current features at different positions in adjacent steps.}
    \label{fig:relation_x_v}
\end{figure*}
\paragraph{The Position to Calculate Similarity}

Figure~\ref{fig:relation_x_v}\textcolor{iccvblue}{a} shows the relationship between the difference of tokens in the adjacent steps at the early layer (3$_{\text{rd}}$ layer) and the output layer. It is observed that \textbf{(I)} The token that exhibits significant changes in the adjacent steps at the early layer is highly likely to demonstrate notable changes at the final layer as well, indicating an approximately linear relationship. This observation explains why LazyMAR can determine whether to cache a token based on the similarity  at the early layer (3$_{\text{rd}}$ layer). \textbf{(II)}
Figure~\ref{fig:relation_x_v}\textcolor{iccvblue}{b-}\ref{fig:relation_x_v}\textcolor{iccvblue}{e} shows the difference of \emph{token features} in the 3$_{\text{rd}}$ layer, the difference of the \emph{value matrix} of tokens in the 3$_{\text{rd}}$ layer, the difference of \emph{token features} in the final layer and the difference of their \emph{value matrix} in the final layer, respectively. The value matrix here indicates the value matrix in self-attention layers.
It is observed that: \emph{The value matrix of tokens are more suitable to be utilized as the criterion to decide whether this token should be computed than the features of tokens.}
As shown in Figure~\ref{fig:relation_x_v}\textcolor{iccvblue}{b}, the tokens decoded in the last step exhibit a clear difference in their features at the 3$_{\text{rd}}$ layer while other tokens show very minor and similar changes. 
In contrast, as shown in Figure~\ref{fig:relation_x_v}\textcolor{iccvblue}{c}, there are more significant differences in the value matrix of the tokens, demonstrating the difference on the value matrix of tokens is more suitable as a criterion to decide whether a token should be computed or reused from the cache. 
\textbf{(III)} Besides, we further compute the Pearson correlation coefficients (PCC) of the difference between tokens in the adjacent steps on the features and value matrix, yielding PCCs of 0.913 and 0.819, respectively, further verifying the previous observation. Hence, we follow it to compute the difference in the value matrix instead of the features to decide whether a token should be computed or reused from the cache.

\begin{figure}
    \centering
    \vspace{-0.1cm}
    \includegraphics[width=\linewidth]{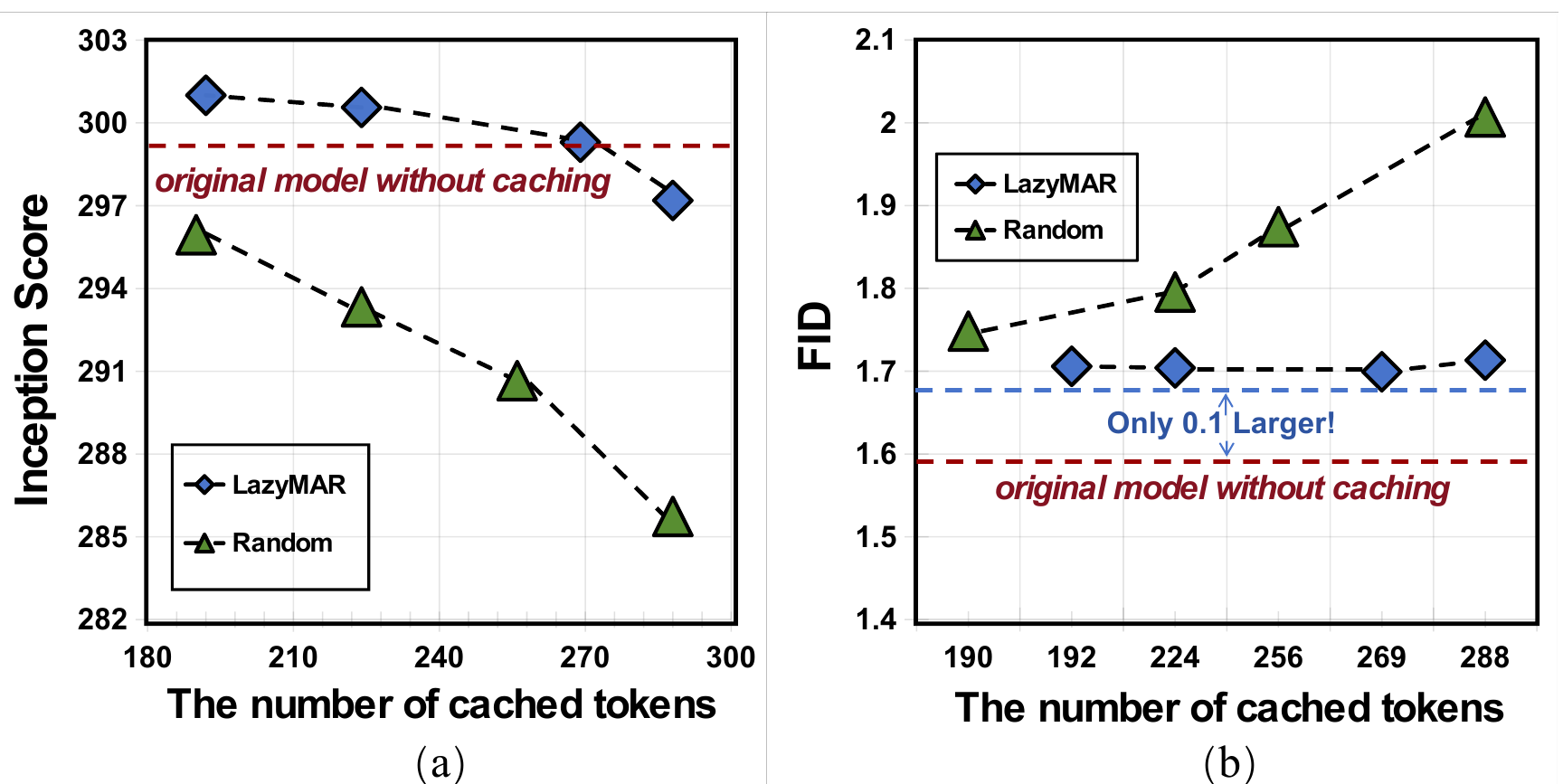}
    \vspace{-0.6cm}    
    \caption{\textbf{Comparison of quality metrics} between LazyMAR and Random Caching under different cache ratios on FID and Inception Scores. As the number of cached tokens increases, the quality metrics of LazyMAR only decrease slightly, while the quality metrics of Random Caching degrade rapidly.}
    \label{fig:pruning_sensible}%
    \vspace{-0.6cm}
\end{figure}

\paragraph{Impact of the Number of Cached Tokens}

At each decoding step, we determine the average number of tokens to cache according to a certain ratio. Figure~\ref{fig:pruning_sensible} shows the comparison between the Token Cache in LazyMAR and Random Cache (\emph{i.e., randomly choosing some tokens for cache reusing},) in FID and Inception Score with different numbers of cached tokens. Compared with Random Cache, LazyMAR maintains better quality metrics when caching the same number of tokens. It is evident that as the number of cached tokens increases, the quality metrics of LazyMAR do not change significantly. In contrast, the quality metrics of the Random Cache suffer a significant decline. It is worth noting that even when caching 90\% (288/320) tokens, LazyMAR can still maintain a comparable FID compared with the original model, demonstrating the advantage of our methods.

\vspace{-0.2cm}
\section{Conclusion}
Motivated by the redundancy in Masked Autoregressive (MAR) models in tokens and classifier-free guidance, 
we introduce LazyMAR, a caching framework that improves the efficiency of MAR with almost no decrease in generation quality. To mitigate token redundancy, we propose Token Cache, which selectively reuses tokens with high similarity to cached values and dynamically computes those with significant changes. By leveraging the correlation between early-layer and final-layer token differences, LazyMAR skips computations effectively without compromising output fidelity.
Additionally, to exploit condition redundancy, we introduce Condition Cache, which caches the residual between conditional and unconditional outputs, eliminating redundant computations in the unconditional branch. Experimental results show that LazyMAR achieves 2.83$\times$ acceleration with minimal impact on FID and Inception Score. Its plug-and-play nature enables seamless integration into existing MAR models without additional training, providing a scalable and practical solution for high-speed masked autoregressive generation.

{\small
    \bibliographystyle{ieeenat_fullname}
    \bibliography{main}
}

\end{document}